\documentclass{article}
\pdfoutput=1

\usepackage[final,nonatbib]{arxiv_2017}

\usepackage[utf8]{inputenc} 
\usepackage[T1]{fontenc}    
\usepackage{hyperref}       
\usepackage{url}            
\usepackage{booktabs}       
\usepackage{amsfonts}       
\usepackage{nicefrac}       
\usepackage{microtype}      
\usepackage{url,amsmath,amssymb,amsthm,braket,algorithmic,algorithm}
\usepackage{graphicx}

\title{Towards Bursting Filter Bubble\\via Contextual Risks and Uncertainties}

\author{
  Rikiya Takahashi\\
  SmartNews, Inc.\\
  6-25-16, Jingumae, Shibuya-ku\\
  Tokyo 150-0001, Japan\\
  \texttt{Rikiya.Takahashi@gmail.com} \\
 \And
  Shunan Zhang\\
  Smartnews International, Inc.\\
  144 2nd St, San Francisco\\
  CA 94105, USA\\
  \texttt{shunan.z@gmail.com} \\
}
\begin{document}

\maketitle

\begin{abstract}
A rising topic in computational journalism is how to enhance the diversity in news served to subscribers to foster exploration behavior in news reading. Despite the success of preference learning in personalized news recommendation, their over-exploitation causes filter bubble that isolates readers from opposing viewpoints and hurts long-term user experiences with lack of serendipity. Since news providers can recommend neither opposite nor diversified opinions if unpopularity of these articles is surely predicted, they can only bet on the articles whose forecasts of click-through rate involve high variability (risks) or high estimation errors (uncertainties). We propose a novel Bayesian model of 
uncertainty-aware scoring and ranking for news articles. The Bayesian binary classifier models probability of success (defined as a news click) as a Beta-distributed random variable conditional on a vector of the context (user features, article features, and other contextual features). The posterior of the contextual coefficients can be computed efficiently using a low-rank version of Laplace’s method via thin Singular Value Decomposition. Efficiencies in personalized targeting of exceptional articles, which are chosen by each subscriber in test period, are evaluated on real-world news datasets. The proposed estimator slightly outperformed existing training and scoring algorithms, in terms of efficiency in identifying successful outliers.
\end{abstract}

\section{Introduction}
\label{sec:intro}

Personalized news recommendation has been a successful application of machine learning
although it would also endanger people's open-mindedness towards opposing viewpoints.
Targeted news delivery has been reinforced by collections of click response logs, 
with preference learning algorithms such as
bilinear models \cite{chu2009personalized,sharma2015feature},
topic models \cite{wu2010topic},
matrix factorization \cite{das2007google,agarwal2010matrix,gunasekar2015consistent}, learning to rank \cite{dali2010learning},
or their hybrid 
\cite{claypool1999combining,zheng2013penetrate}.
In terms of exploration-exploitation trade-off, however, the exploitative nature of data-oriented personalization
has been criticized as a cause of the filter bubble \cite{pariser2011filter},
which is a compound problem fueled both by narrow-minded human's willing choice \cite{bakshy2015exposure}
and by machines over-confident about their scoring.
In this paper, we address the problem of over-confident machines while psychologically opening people's mind is out of the current scope.

Even by diversifying recommendations or by exploring around uncertain preferences,
online news providers cannot deliver articles whose unpopularity is surely predicted.
Submodular maximization algorithms \cite{yue2011linear} or 
Determinantal Point Processes (DPPs \cite{affandi2012markov,gillenwater2014em,ren2015summarizing}; see \cite{kulesza2013determinantal} for good summary)
generate a subset of input items associated with univariate scores (e.g., estimated click-through rates).
The output subset consists of some high-score items that are well diversified.
Unfortunately, due to the exclusion of low-score items,
diversification algorithms never select items whose unpopularity is statistically significant.
Another direction towards serendipity is to use not unbiased but optimistically-biased estimates
as in contextual multi-armed bandit algorithms
\cite{liu2010personalized,li2010contextual,yue2011linear,may2012optimistic,tang2014ensemble}.
Bandit algorithms, however, also fail to select surely unpopular items,
because expected cumulative reward is their main objective in repeated trials.
As a consequence, as long as a large number of total clicks must be retained,
news providers cannot bet against readers' preference evidenced by large samples.

We instead recommend articles whose popularity involves high variabilities (risks) and/or high estimation errors (uncertainties),
by introducing a new Bayesian classifier that explicitly formalizes the dependence of variabilities on recommendation contexts.
It is worth observing that Maximum A Posteriori (MAP) point estimation and logistic-loss models (e.g., factorization machines
\cite{rendle2012factorization,juan2016field} and deep neural networks \cite{covington2016deep}) are broadly used in real-world systems.
We hypothesize that content providers too much rely on the logistic-loss MAP estimates 
because of their successes in retrieving popular items, 
and they often forget the fact that substance of their model
is an erroneous estimate deviated from the true preference.
Furthermore, we gaze at a statistical insight that estimation error of variability statistic is particularly high when the true variability is high.
By performing Bayesian interval estimation and assuming the true variability to be a function of context,
we robustly quantify the total stochasticity with clear separation between the essential variability and estimation error.
These deliberated philosophies behind our model are expected to increase the chance of discovering exceptional items,
whose preference uncertainty has been underestimated by the existing models.

Our key ideas to accurately quantify the context-dependent risks and uncertainties 
are conditional Beta distribution and low-rank Laplace approximation with affordable computational costs.
We model the probability of success as  a random variable to obey a Beta distribution,
whose two parameters are both functions of input vector.
In training, we approximate the posterior by a low-rank Gaussian distribution
whose estimates of the variance-covariance matrices are given by thin Singular Value Decomposition (SVD).
Thanks to this thin SVD,
we can perform both training and recommendation
efficiently by sparse high-dimensional matrix libraries.
We also derive a variety of uncertainty-aware scoring measures
over the closed forms of the approximate posterior, and evaluate the performance of every measure by using real-world news-reading datasets.
While we compare linear models in the experiments,
since our model is a robust extension of logistic regression with additional dispersion parameters,
the principle of combining conditional Beta distribution with low-rank Laplace's method
is broadly pluggable into many logistic-loss non-linear models.

The remainder of this paper is organized as follows.
Our Bayesian binary classifier and its approximate posterior inference are introduced in Section \ref{sec:training}.
We then derive several types of posterior-based scoring methods in Section \ref{sec:recommendation}.
Section \ref{sec:related_work} discusses the related work on robust classification
and human's exploration behavior though the latter theme is out of our scope.
In Section \ref{sec:experiments}, we numerically evaluate the proposed algorithm and scoring measures
for the news datasets, with introducing a novel performance indicator for serendipity-oriented targeting.
Section \ref{sec:conclusion} concludes the paper.

\section{Bayesian training of a binary classifier with contextual variabilities}
\label{sec:training}

Let us introduce a Bayesian binary classifier in which
the probability of success  is a random variable conditional on vector of context.
As shown in Section \ref{subsec:beta_binomial_model},
the frequency of success obeys a Beta-binomial distribution
and we place a Gaussian prior.
Though the exact posterior is intractable, 
low-rank Laplace's method introduced in Section \ref{subsec:laplace_approximation}
provides a closed-form estimate with affordable costs in matrix computation,
and also leads derivation of an approximate marginal likelihood in Section \ref{subsec:hyperparameter_section}.
The low-rank approximate posterior on the maximum-marginal-likelihood prior hyperparameters
is finally used for recommendation whose details are later provided in Section \ref{sec:recommendation}.

\subsection{Beta-binomial model and Gaussian prior}
\label{subsec:beta_binomial_model}

Let $\theta_{ij}\!\in\![0, 1]$ be probability with which user $i$ chooses item $j$, and 
let $\boldsymbol{x}_{ij} \in {\mathbb R}^d$ be vector of context 
when item $j$ is shown to user $i$.
In our news example, each item is an article and the vector $\boldsymbol{x}_{ij}$
is defined on a conjunction of news subscriber $i$'s characteristics and text content of article $j$.
We assume that probability $\theta_{ij}$ 
obeys a conditional Beta distribution on vector $\boldsymbol{x}_{ij}$ as
\begin{equation}
\label{equation:beta_model}
p(\theta_{ij}\vert\boldsymbol{x}_{ij}, \boldsymbol{w})\!=\!Be\!\left(\theta_{ij}; %
\exp((\boldsymbol{\beta}+\boldsymbol{\rho})^\top\boldsymbol{x}_{ij}), %
\exp(\boldsymbol{\rho}^\top\boldsymbol{x}_{ij})\right)\!,
\end{equation}
where $Be\!\left(\theta; \alpha_1, \alpha_2\right)\!\triangleq\!%
\frac{\Gamma(\alpha_1\!+\!\alpha_2)}{\Gamma(\alpha_1)\Gamma(\alpha_2)}%
\theta^{\alpha_1-1}(1\!-\!\theta)^{\alpha_2-1}$
and $\boldsymbol{w}\!\triangleq\!(\boldsymbol{\beta}^\top,\boldsymbol{\rho}^\top)^\top$
is a vector of regression coefficients. Let us define expectation operator ${\mathbb E}_{p}$ as 
${\mathbb E}_{p(\theta)}[f(\theta)]\!\triangleq\!\int_{\theta}f(\theta)p(\theta)d\theta$. 
Since ${\mathbb E}_{p(\theta_{ij}\vert\boldsymbol{x}_{ij}, \boldsymbol{w})}[\theta_{ij}]\!=\!1/(1+\exp(-\boldsymbol{\beta}^\top\boldsymbol{x}_{ij}))$,
Eq.~(\ref{equation:beta_model}) is interpreted as a logistic regression
with additional variabilities introduced by unobservable variables that are not contained in vector $\boldsymbol{x}_{ij}$.

We estimate posterior of $\boldsymbol{w}$.
Let ${\mathcal N}(\cdot; \boldsymbol{\mu}, \boldsymbol{\Sigma})$
be the probability density function of multivariate Gaussian distribution
whose mean is $\boldsymbol{\mu}$ and whose variance-covariance matrix 
is $\boldsymbol{\Sigma}$.
We place an isotropic Gaussian prior $p(\boldsymbol{w})={\mathcal N}(\boldsymbol{\beta}; \boldsymbol{0}_{d}, c_{\boldsymbol{\beta}}^{-1}\boldsymbol{I}_{d}){\mathcal N}(\boldsymbol{\rho}; \boldsymbol{0}_{d}, c_{\boldsymbol{\rho}}^{-1}\boldsymbol{I}_{d})$ where $\boldsymbol{0}_d$ and $\boldsymbol{I}_d$ are the $d$-dimensional zero vector and identity matrix, and $\boldsymbol{c}\!\triangleq\!(c_{\boldsymbol{\beta}}, c_{\boldsymbol{\rho}})^\top$ is a vector of prior hyperparameters.

Because we observe only choice frequencies,
the data likelihood is given by probability mass function of Beta-binomial distribution.
Let $n_{ij}$ and $v_{ij}$ be the frequencies with which user $i$ is exposed to item $j$
and with which user $i$ chooses item $j$, respectively. 
In online news recommendation, $n_{ij}$ and $v_{ij}$ are called 
the numbers of impressions and page views, respectively.
Since $v_{ij}$ is distributed from binomial distribution whose number of trials is $n_{ij}$
and whose probability of success is $\theta_{ij}$, we can obtain the data likelihood 
marginal over the latent choice probabilities as
\begin{equation}
p(v_{ij}\vert n_{ij},\boldsymbol{x}_{ij},\boldsymbol{w})\!\triangleq\!%
{\mathbb E}_{p(\theta_{ij}\vert\boldsymbol{x}_{ij}, \boldsymbol{w})}\!\!\left[\theta_{ij}^{v_{ij}}\!(1\!-\!\theta_{ij})^{n_{ij}\!-\!v_{ij}}\!\right]%
\!\!=\!BB\!\left(v_{ij}; n_{ij},\exp(\!(\boldsymbol{\beta}\!+\!\boldsymbol{\rho})\!^\top\!\boldsymbol{x}_{ij}\!),%
\exp(\boldsymbol{\rho}\!^\top\!\!\boldsymbol{x}_{ij}\!)\right)\!\nonumber,
\end{equation}
where $BB\!\left(v; n, \alpha_1, \alpha_2\right)$ represents the
unnormalized probability mass function of Beta-binomial distribution
such that $BB\!\left(v; n, \alpha_1, \alpha_2\right)\!\triangleq\!%
\frac{\Gamma(\alpha_1\!+\!\alpha_2)}{\Gamma(\alpha_1\!+\!\alpha_2\!+\!n)}%
\frac{\Gamma(\alpha_1\!+\!v)}{\Gamma(\alpha_1)}%
\frac{\Gamma(\alpha_2\!+\!n\!-\!v)}{\Gamma(\alpha_2)}$.
Since Beta-binomial and Gaussian distributions are not conjugate, the true posterior is analytically intractable.

\subsection{Block low-rank Laplace approximation}
\label{subsec:laplace_approximation}

Laplace's method is useful in approximate Bayesian inference particularly when the true posterior is
close to a Gaussian distribution. The approximate Gaussian posterior's 
mean is given by the Maximum A Posteriori (MAP) estimate and 
its variance-covariance matrix is obtained through a quadratic approximation around the MAP estimate. 
Our negative joint log-likelihood is given as ${\mathcal L}_{{\mathcal D},\boldsymbol{c}}(\boldsymbol{w})\!\triangleq\!%
-\sum_{(i,j)\in{\mathcal D}} \log BB\left(v_{ij}; n_{ij}, \exp((\boldsymbol{\beta}+\boldsymbol{\rho})^\top\boldsymbol{x}_{ij}),%
\exp(\boldsymbol{\rho}^\top\boldsymbol{x}_{ij})\right)%
+\frac{c_{\boldsymbol{\beta}}}{2}\Vert\boldsymbol{\beta}\Vert^2%
+\frac{c_{\boldsymbol{\rho}}}{2}\Vert\boldsymbol{\rho}\Vert^2$,
where  $\Vert\cdot\Vert$ denotes the $L_2$ norm
and ${\mathcal D}=\lbrace (i,j) \rbrace$ is the training set of user-item pairs.
Let
\begin{eqnarray}
\alpha_{ij}^{+}&\!\triangleq\!&\exp((\boldsymbol{\beta}\!+\!\boldsymbol{\rho})\!^\top\!\boldsymbol{x}_{ij}),~%
\alpha_{ij}^{-}\!\triangleq\!\exp(\boldsymbol{\rho}\!^\top\!\boldsymbol{x}_{ij}),~%
\alpha_{ij}\!\triangleq\!\alpha_{ij}^{+}+\alpha_{ij}^{-},\nonumber\\
g_{ij}^{(\boldsymbol{\beta})}&\!\triangleq\!&
-\alpha_{ij}^{+}\left[\Psi(\alpha_{ij}^{+}+v_{ij})\!-\!\Psi(\alpha_{ij}^{+})%
\!-\!\Psi(\alpha_{ij}+n_{ij})\!+\!\Psi(\alpha_{ij})\right]\nonumber,\\
g_{ij}^{(\boldsymbol{\rho})}&\!\triangleq\!&%
-\alpha_{ij}^{+}\left[\Psi(\alpha_{ij}^{+}+v_{ij})\!-\!\Psi(\alpha_{ij}^{+})\right]-%
\alpha_{ij}^{-}\left[\Psi(\alpha_{ij}^{-}+n_{ij}-v_{ij})\!-\!\Psi(\alpha_{ij}^{-})\right]%
\nonumber\\
&&+\alpha_{ij}\left[\Psi(\alpha_{ij}+n_{ij})\!-\!\Psi(\alpha_{ij})\right]\nonumber,\\
h_{ij}^{(\boldsymbol{\beta})}&\!\triangleq\!&%
-(\alpha_{ij}^{+})^2\left[\Psi'(\alpha_{ij}^{+}+v_{ij})\!-\!\Psi'(\alpha_{ij}^{+})%
\!-\!\Psi'(\alpha_{ij}+n_{ij})\!+\!\Psi'(\alpha_{ij})\right]+g_{ij}^{(\boldsymbol{\beta})}\nonumber,\\
h_{ij}^{(\boldsymbol{\beta},\boldsymbol{\rho})}&\!\triangleq\!&%
-(\alpha_{ij}^{+})^2\left[\Psi'(\alpha_{ij}^{+}+v_{ij})\!-\!\Psi'(\alpha_{ij}^{+})\right]%
+\alpha_{ij}\alpha_{ij}^{+}\left[\Psi'(\alpha_{ij}+n_{ij})\!-\!\Psi'(\alpha_{ij})\right]+g_{ij}^{(\boldsymbol{\beta})}\nonumber\mbox{, and}\\
h_{ij}^{(\boldsymbol{\beta})}&\!\triangleq\!&%
-(\alpha_{ij}^{+})^2\left[\Psi'(\alpha_{ij}^{+}+v_{ij})\!-\!\Psi'(\alpha_{ij}^{+})\right]%
-(\alpha_{ij}^{-})^2\left[\Psi'(\alpha_{ij}^{-}+n_{ij}-v_{ij})\!-\!\Psi'(\alpha_{ij}^{-})\right]\nonumber\\
&&+\alpha_{ij}^2\left[\Psi'(\alpha_{ij}+n_{ij})\!-\!\Psi'(\alpha_{ij})\right]+g_{ij}^{(\boldsymbol{\rho})}\nonumber,
\end{eqnarray}
where $\Psi(\cdot)$ and $\Psi'(\cdot)$ are the digamma and trigamma functions such that 
$\Psi(u)\!\triangleq\!\frac{\partial\log\Gamma(u)}{\partial u}$ and 
$\Psi'(u)\!\triangleq\!\frac{\partial\Psi(u)}{\partial u}$.
The MAP estimate 
$\widehat{\boldsymbol{w}}\!=\!\arg\min_{\boldsymbol{w}}{\mathcal L}_{{\mathcal D},\boldsymbol{c}}(\boldsymbol{w})$
is attained by descending the gradient 
$\frac{\partial{\mathcal L}_{{\mathcal D},\boldsymbol{c}}(\boldsymbol{w})}{\partial\boldsymbol{w}}%
\!=\!%
\sum_{(i,j)\in{\mathcal D}}%
(g_{ij}^{(\boldsymbol{\beta})}\boldsymbol{x}_{ij}^\top,%
g_{ij}^{(\boldsymbol{\rho})}\boldsymbol{x}_{ij}^\top)^\top%
\!+\!%
(c_{\boldsymbol{\beta}}\boldsymbol{\beta}^\top,%
c_{\boldsymbol{\rho}}\boldsymbol{\rho}^\top)^\top$
with the Hessian matrix
\begin{equation}
\label{equation:data_hessian}
\dfrac{\partial^2{\mathcal L}_{{\mathcal D},\boldsymbol{c}}(\boldsymbol{w})}{\partial\boldsymbol{w}\partial\boldsymbol{w}^\top}%
~=\!\sum_{(i,j)\in{\mathcal D}}%
\!\begin{pmatrix}%
h_{ij}^{(\boldsymbol{\beta})}\boldsymbol{x}_{ij}\boldsymbol{x}_{ij}^\top &
h_{ij}^{(\boldsymbol{\beta},\boldsymbol{\rho})}\boldsymbol{x}_{ij}\boldsymbol{x}_{ij}^\top \\
h_{ij}^{(\boldsymbol{\beta},\boldsymbol{\rho})}\boldsymbol{x}_{ij}\boldsymbol{x}_{ij}^\top &
h_{ij}^{(\boldsymbol{\rho})}\boldsymbol{x}_{ij}\boldsymbol{x}_{ij}^\top
\end{pmatrix}+%
\begin{pmatrix}%
c_{\boldsymbol\beta}\boldsymbol{I}_d & \boldsymbol{O}_d\\
\boldsymbol{O}_d & c_{\boldsymbol\rho}\boldsymbol{I}_d
\end{pmatrix},%
\end{equation}
where $\boldsymbol{O}_d$ is the $d$-by-$d$ zero matrix.
After convergence,
we approximate the posterior by a factorial form
$q(\boldsymbol{w}\vert {\mathcal D},\boldsymbol{c})\!\triangleq\!%
{\mathcal N}\!\left(\!\boldsymbol{w}; \widehat{\boldsymbol{w}},\widehat{\boldsymbol{\Sigma}}_{\boldsymbol{w}}\!\right)\!\equiv\!%
{\mathcal N}\!\left(\!\boldsymbol{\beta}; \widehat{\boldsymbol{\beta}},\widehat{\boldsymbol{\Sigma}}_{\boldsymbol{\beta}}\!\right)%
{\mathcal N}\!\left(\!\boldsymbol{\rho}; \widehat{\boldsymbol{\rho}},\widehat{\boldsymbol{\Sigma}}_{\boldsymbol{\rho}}\!\right)$%
\footnote{
We avoid to parametrize $\widehat{\boldsymbol{\Sigma}}_{\boldsymbol{w}}$ by authentic full variance-covariance matrix in this paper,
due to the too lengthy closed-form expression stemming from the complex correlation between $\boldsymbol{\beta}$ and $\boldsymbol{\rho}$.}.

The high-dimensionality of context vector $\boldsymbol{x}_{ij}$, however,
does not allow us for materializing each block's full variance-covariance matrix,
whose further diagonalization is neither acceptable due to the poor quantification of uncertainty
when ignoring multi-collinearity among different words in text.

We instead estimate a rank-$k$ ($\ll\!d$) approximation of the inverse variance-covariance matrix as
\begin{equation}
\label{equation:approximate_inverse_covmatrix}
\widehat{\boldsymbol{\Sigma}}_{\boldsymbol{\beta}}^{-1}=%
\boldsymbol{V}_{\boldsymbol{\beta}}%
\boldsymbol{\Lambda}_{\boldsymbol{\beta}}%
\boldsymbol{V}_{\boldsymbol{\beta}}^\top%
+c_{\boldsymbol{\beta}}\boldsymbol{I}_d\mbox{ and }%
\widehat{\boldsymbol{\Sigma}}_{\boldsymbol{\rho}}^{-1}=%
\boldsymbol{V}_{\boldsymbol{\rho}}%
\boldsymbol{\Lambda}_{\boldsymbol{\rho}}%
\boldsymbol{V}_{\boldsymbol{\rho}}^\top%
+c_{\boldsymbol{\rho}}\boldsymbol{I}_d,%
\end{equation}
where 
$\boldsymbol{V}_{\boldsymbol{\beta}},\boldsymbol{V}_{\boldsymbol{\rho}}\!\in\!{\mathbb R}^{d\times k}$,
$\boldsymbol{\Lambda}_{\boldsymbol{\beta}}\!\equiv\!diag(\lambda_{\boldsymbol{\beta} 1},\ldots,\lambda_{\boldsymbol{\beta} k})$, and
$\boldsymbol{\Lambda}_{\boldsymbol{\rho}}\!\equiv\!diag(\lambda_{\boldsymbol{\rho} 1},\ldots,\lambda_{\boldsymbol{\rho} k})$.
By carefully watching Eq.~(\ref{equation:data_hessian}),
one can find that each matrix $\boldsymbol{V}_{(\cdot)}$ is obtained by
thin Singular Value Decomposition (SVD).
Let us assume that users and items are indexed from $i\!=\!1$ to $i\!=\!I$
and from $j\!=\!1$ to $j\!=\!J$, respectively.
For each block, we perform thin SVD of a weighted data matrix as
\begin{equation}
\boldsymbol{X}_{(\cdot)}\!\triangleq\!%
(({h_{11}^{(\cdot)}\vert_{\boldsymbol{w}\!=\!\widehat{\boldsymbol{w}}}})^{1/2}\boldsymbol{x}_{11},%
\ldots,%
({h_{IJ}^{(\cdot)}\vert_{\boldsymbol{w}\!=\!\widehat{\boldsymbol{w}}}})^{1/2}\boldsymbol{x}_{IJ})%
^{\!\!\top}%
\!\!\in{\mathbb R}^{\vert {\mathcal D}\vert\times d}
\mbox{ and }
\boldsymbol{X}_{(\cdot)}\simeq\boldsymbol{U}_{(\cdot)}%
\boldsymbol{\Lambda}_{(\cdot)}^{\frac{1}{2}}%
\boldsymbol{V}_{(\cdot)}^\top\nonumber,
\end{equation}
where $\boldsymbol{U}_{(\cdot)}\!\in\!{\mathbb R}^{\vert{\mathcal D}\vert\times k}$
and matrix $\boldsymbol{\Lambda}_{(\cdot)}^{\frac{1}{2}}$
is given by the $k$-largest singular values of $\boldsymbol{X}_{(\cdot)}$.
Based on the rank-$k$ approximation,
the closed form of each variance-covariance matrix is consequently given as  
\begin{equation}
\label{equation:closed_form_covariance}
\widehat{\boldsymbol{\Sigma}}_{(\cdot)}\!\triangleq\!\left(c_{(\cdot)}\boldsymbol{I}_d\!+\!\boldsymbol{V}_{(\cdot)}\boldsymbol{\Lambda}_{(\cdot)}\boldsymbol{V}_{(\cdot)}^\top\right)^{-1}\!\equiv\!%
c_{(\cdot)}^{-1}\boldsymbol{I}_d\!-\!c_{(\cdot)}^{-1}\boldsymbol{V}_{(\cdot)}(\boldsymbol{I}_d+c\boldsymbol{\Lambda}_{(\cdot)}^{-1})^{-1}\boldsymbol{V}_{(\cdot)}^\top.
\end{equation}

\subsection{Hyperparameter optimization with approximate marginal likelihood}
\label{subsec:hyperparameter_section}

Laplace's method also provides a closed-form approximation of the marginal likelihood and 
enables to optimize the hyperparameter vector $\boldsymbol{c}$ in a Bayesian manner.
Our loss is quadratically approximated around the MAP estimate $\widehat{\boldsymbol{w}}$ as
${\mathcal L}_{{\mathcal D},\boldsymbol{c}}(\boldsymbol{w})\!\simeq\!%
{\mathcal L}_{{\mathcal D},\boldsymbol{c}}(\widehat{\boldsymbol{w}})\!+\!%
\frac{1}{2}(\boldsymbol{w}\!-\!\widehat{\boldsymbol{w}})^\top%
\widehat{\boldsymbol{\Sigma}}_{\boldsymbol{w}}^{-1}(\boldsymbol{w}\!-\!\widehat{\boldsymbol{w}})$.
The marginal negative log-likelihood
$\widetilde{{\mathcal L}}_{{\mathcal D},\boldsymbol{c}}\!\triangleq\!%
-\log\int d\boldsymbol{w}\prod_{(i,j)\in{\mathcal D}}p(v_{ij}\vert n_{ij}, \boldsymbol{x}_{ij}, \boldsymbol{w})p(\boldsymbol{w})$ is hence approximated as
\begin{eqnarray}
\exp(-\widetilde{{\mathcal L}}_{{\mathcal D},\boldsymbol{c}})\!&\simeq&\!%
p(\widehat{\boldsymbol{w}})\!\!\prod_{(i,j)\in{\mathcal D}}\!\!\!p(v_{ij}\vert n_{ij}, \boldsymbol{x}_{ij}, \widehat{\boldsymbol{w}})%
\int\exp\left(-\frac{1}{2}(\boldsymbol{w}\!-\!\widehat{\boldsymbol{w}})^\top%
\widehat{\boldsymbol{\Sigma}}_{\boldsymbol{w}}^{-1}(\boldsymbol{w}\!-\!\widehat{\boldsymbol{w}})\right)d\boldsymbol{w}%
\nonumber\\
\!&\equiv&\!
\exp\left(-{\mathcal L}_{{\mathcal D},\boldsymbol{c}}(\widehat{\boldsymbol{w}})\right)
\vert\widehat{\boldsymbol{\Sigma}}_{\boldsymbol{\beta}}\vert^{1/2}
\vert\widehat{\boldsymbol{\Sigma}}_{\boldsymbol{\rho}}\vert^{1/2}
c_{\boldsymbol{\beta}}^{d/2}c_{\boldsymbol{\rho}}^{d/2}\nonumber.
\end{eqnarray}
Because each determinant is the product of diagonal elements of the matrix $\boldsymbol{\Lambda}_{(\cdot)}$ in (\ref{equation:approximate_inverse_covmatrix}),
the approximate marginal negative log-likelihood to select the hyperparameter $\boldsymbol{c}$ is 
\begin{equation}
\label{equation:marginal_loss}
\widetilde{{\mathcal L}}_{{\mathcal D},\boldsymbol{c}}
\simeq
{\mathcal L}_{{\mathcal D},\boldsymbol{c}}(\widehat{\boldsymbol{w}})
+\frac{1}{2}\sum_{\ell=1}^k
\log\left(1+\frac{\lambda_{\boldsymbol{\beta}\ell}}{c_{\boldsymbol{\beta}}}\right)%
\!+\frac{1}{2}\sum_{\ell=1}^k\log\left(1+\frac{\lambda_{\boldsymbol{\rho}\ell}}{c_{\boldsymbol{\rho}}}\right)\nonumber.
\end{equation}
Based on the $L_2$-regularization terms in the loss function ${\mathcal L}_{{\mathcal D},\boldsymbol{c}}$ and Eq.~(\ref{equation:marginal_loss}),
we maximize the marginal likelihood by iterating between 
the update of $(\widehat{\boldsymbol{w}}, \widehat{\boldsymbol{\Sigma}}_{\boldsymbol{w}})$ and that of $\boldsymbol{c}$ as
\begin{equation}
c_{\boldsymbol{\beta}}\!=\!\min_{c}\left[c\Vert\widehat{\boldsymbol{\beta}}\Vert^2\!+\!\sum_{\ell=1}^k\log\left(1\!+\!\frac{\lambda_{\boldsymbol{\beta}\ell}}{c}\right)\right]\mbox{ and }%
c_{\boldsymbol{\rho}}\!=\!\min_{c}\left[c\Vert\widehat{\boldsymbol{\rho}}\Vert^2\!+\!\sum_{\ell=1}^k\log\left(1\!+\!\frac{\lambda_{\boldsymbol{\rho}\ell}}{c}\right)\right]\nonumber.
\end{equation}

\section{Uncertainty-aware recommendation}
\label{sec:recommendation}

In order to rank test items, this section introduces several scoring measures 
which are all derived from the approximate posterior or predictive distribution.
While the exact predictive distribution is intractable
due to the non-conjugacy between Beta and Gaussian distributions,
its Monte Carlo approximation is easily obtained as we show 
in Section \ref{subsec:monte_carlo_predictive}. The Monte Carlo approach
eases computation of the scoring measures, whose varieties and characteristics are described in
\ref{subsec:varieties_scores}. 
Dependence of recommendation results on the choice of scoring methods is evaluated in Section \ref{sec:experiments}. 

\subsection{Monte-Carlo predictive distribution}
\label{subsec:monte_carlo_predictive}

Let $\theta_{*}\!\in\![0,1]$  be probability of success in test context associated with
vector $\boldsymbol{x}_{*}\!\in\!{\mathbb R}^d$.
Integrating over the approximate posterior of $\boldsymbol{w}$,
we obtain the predictive distribution of $\theta_{*}$ conditional on $\boldsymbol{x}_{*}$.
The approximate predictive distribution is defined as
$q(\theta_{*}\vert\boldsymbol{x}_{*}, {\mathcal D},\boldsymbol{c})\!\triangleq\!%
{\mathbb E}_{q(\boldsymbol{w}\vert{\mathcal D},\boldsymbol{c})}\left[p(\theta_{*}\vert\boldsymbol{x}_{*}, \boldsymbol{w})\right]\!=\!%
{\mathbb E}_{{\mathcal N}\left(\boldsymbol{\beta}; \widehat{\boldsymbol{\beta}},\widehat{\boldsymbol{\Sigma}}_{\boldsymbol{\beta}}\right)%
{\mathcal N}\left(\boldsymbol{\rho}; \widehat{\boldsymbol{\rho}},\widehat{\boldsymbol{\Sigma}}_{\boldsymbol{\rho}}\right)}
\!\left[Be\!\left(\theta_{*}; %
\exp((\boldsymbol{\beta}\!+\!\boldsymbol{\rho})\!^\top\!\boldsymbol{x}_{*}), %
\exp(\boldsymbol{\rho}\!^\top\!\boldsymbol{x}_{*})\right)\right]$, whose inherent integral is analytically intractable
while is numerically well-approximated by a bidimensional Monte Carlo integration.
Since each block of the variance-covariance matrix
has the common form (\ref{equation:closed_form_covariance}),
variances of $\zeta_{*}\!\triangleq\!\boldsymbol{\beta}^\top\boldsymbol{x}_{*}$ and $\eta_{*}\!\triangleq\!\boldsymbol{\rho}^\top\boldsymbol{x}_{*}$ are
both given as
$\boldsymbol{x}_{*}^\top\widehat{\boldsymbol{\Sigma}}_{(\cdot)}\boldsymbol{x}_{*}\!\equiv\!c_{(\cdot)}^{-1}\boldsymbol{x}_{*}^\top\boldsymbol{x}_{*}\!-\!c_{(\cdot)}^{-1}\boldsymbol{x}_{*}^\top\boldsymbol{V}_{(\cdot)}(\boldsymbol{I}_d+c_{(\cdot)}\boldsymbol{\Lambda}_{(\cdot)}^{-1})^{-1}\boldsymbol{V}_{(\cdot)}^\top\boldsymbol{x}_{*}$. Therefore, an $m$-sample Monte Carlo approximation of the predictive distribution is given as
\begin{eqnarray}
\sigma_{(\cdot)}^2(\boldsymbol{x}_{*})\!\!\!\!&\triangleq&\!\!\!\!%
c_{(\cdot)}^{-1}\!\left(\Vert\boldsymbol{x}_{*}\Vert^2%
\!-\!\sum_{\ell=1}^k \frac{\lambda_{(\cdot)\ell}}{\lambda_{(\cdot)\ell}+c_{(\cdot)}}y_{*(\cdot)\ell}^2\right)%
\mbox{where }(y_{(\cdot)*1},\ldots,y_{(\cdot)*k})^\top\!\triangleq\!\boldsymbol{V}_{(\cdot)}^\top\boldsymbol{x}_{*}
\nonumber,\\
\label{equation:monte_carlo_scores}
\begin{pmatrix}%
\zeta_{*}^{(1)}\\
\eta_{*}^{(1)}%
\end{pmatrix}%
\!,\ldots\!,%
\begin{pmatrix}%
\zeta_{*}^{(m)}\\
\eta_{*}^{(m)}%
\end{pmatrix}%
\!\!\!\!&\sim&\!\!\!\!%
{\mathcal N}\!\left(%
\!\!\begin{pmatrix}%
\widehat{\boldsymbol{\beta}}^\top\boldsymbol{x}_{*}\\
\widehat{\boldsymbol{\rho}}^\top\boldsymbol{x}_{*}
\end{pmatrix}\!, %
\begin{pmatrix}%
\sigma_{\boldsymbol{\beta}}^2(\boldsymbol{x}_{*}) & 0\\
0 & \sigma_{\boldsymbol{\rho}}^2(\boldsymbol{x}_{*})
\end{pmatrix}\right)\mbox{, and }\\
\label{equation:approximate_predictive_distribution}
q(\theta_{*}\vert\boldsymbol{x}_{*}, {\mathcal D},\boldsymbol{c})\!\!\!\!&\simeq&\!\!\!\!%
\frac{1}{m}\sum_{t=1}^m
Be\!\left(\theta_{*}; %
\exp(\zeta_{*}^{(t)}+\eta_{*}^{(t)}), %
\exp(\zeta_{*}^{(t)})\right).%
\end{eqnarray}
The total complexity in (\ref{equation:monte_carlo_scores}) is ${\mathcal O}(\max\lbrace dk, m\rbrace)$ while it is
much lower for sparse $\boldsymbol{x}_{*}$.
Thanks to the low-dimensionality of the integral,
Quasi-Monte Carlo method makes the integration more accurate.

\subsection{Varieties of recommendation scores}
\label{subsec:varieties_scores}

By using Eqs.~(\ref{equation:monte_carlo_scores}) and (\ref{equation:approximate_predictive_distribution}), we can
derive several types of scores used in the final recommendation.
With regarding probability of success as our target variable, 
we introduce a variety of measures based on Upper Confidence Bounds (UCBs) or quantiles. While
many of our measures stem from existing approaches to handle exploration-exploitation trade-off in multi-armed bandit or Bayesian optimization,
our recommendation experiments in Section \ref{sec:experiments} are not repetitious but one-time.

\paragraph{Upper Confidence Bound of Expectation (UCBE)}

UCB in multi-armed bandit \cite{yue2011linear} or Bayesian optimization 
\cite{contal2014gaussian} is usually defined as
an optimistic estimate of the expected reward under uncertainty. In our model,
the expected reward under no uncertainty of $\boldsymbol{w}$
is ${\mathbb E}_{p(\theta_{*}\vert\boldsymbol{x}_{*}, \boldsymbol{w})}[\theta_{*}]\!=\!1/(1+\exp(-\boldsymbol{\beta}^\top\!\boldsymbol{x}_{*}))$, 
which does not depend on $\boldsymbol{\rho}$.
By using the Monte Carlo samples $\zeta_{*}^{(1)},\ldots,\zeta_{*}^{(m)}$ in (\ref{equation:monte_carlo_scores}),
we obtain the mean and standard deviation about the probability of success as
\begin{equation}
\mu_{*}\!\simeq\!\frac{1}{m}\sum_{t=1}^m \frac{1}{1+\exp\left(-\zeta_{*}^{(t)}\right)}\mbox{ and }%
s_{*}^2\!\simeq\!\frac{1}{m\!-\!1}\sum_{t=1}^m \left(\frac{1}{(1+\exp\left(-\zeta_{*}^{(t)}\right)}\!-\!\mu_{*}\right)^2\nonumber,
\end{equation}
respectively. Then the $\nu$-UCBE is finally computed as
\begin{equation}
\label{equation:our_ucb}
{UCBE}_{q(\theta_{*}\vert\boldsymbol{x}_{*}, {\mathcal D},\boldsymbol{c})}^{(\nu)}[\theta_{*}]\!\simeq\!%
\mu_{*}+\Phi^{-1}(\nu)s_{*}\nonumber,
\end{equation}
where $\Phi^{-1}(\cdot)$ is the inverse cumulative distribution function of the standard normal distribution
and $\nu\!>\!0.5$ is a hyperparameter to determine the exploration-exploitation trade-off.
While Eq.~(\ref{equation:our_ucb}) does not explicitly depend on $\boldsymbol{\rho}$, 
the additional complexity by $\boldsymbol{\rho}$ affects the estimate of $\boldsymbol{\beta}$.
The explicit handling of over-dispersion more robustifies UCB than the standard logistic regression.

\paragraph{Upper Confidence Quantile of Expectation (UCQE)}

Another optimistic statistic easily obtained from the posterior is upper quantile of 
the expected reward. Specifically, we compute $100\nu$-percentile 
of ${\mathbb E}_{p(\theta_{*}\vert\boldsymbol{x}_{*}, \boldsymbol{w})}\left[\theta_{*}\right]$ over the posterior 
$q(\theta_{*}\vert\boldsymbol{x}_{*}, {\mathcal D},\boldsymbol{c})$. Because the sigmoid function is a monotonic transformation, 
quantile of the sigmoid is the sigmoid of the quantile, whose computation does not require Monte Carlo samples.
Hence our $\nu$-UCQE measure is given as
\begin{equation}
{UCQE}_{q(\theta_{*}\vert\boldsymbol{x}_{*}, {\mathcal D},\boldsymbol{c})}^{(\nu)}[\theta_{*}]\!=\!%
\frac{1}{1+\exp(-\zeta_{*\nu})}\mbox{ where }
\zeta_{*\nu}=\widehat{\boldsymbol{\beta}}^\top\boldsymbol{x}_{*}+\Phi^{-1}(\nu)\sigma_{\boldsymbol{\beta}}(\boldsymbol{x}_{*})\nonumber.
\end{equation}
Also for UCQE, modeling the over-dispersion introduced by $\boldsymbol{\rho}$
supplies robustification.

\paragraph{Expected Upper Quantile (EUQ)}

Upper quantile of the reward is a valuable statistic when we are interested in outliers
whose probability $\theta_{*}$
is not perfectly predictable from context vector $\boldsymbol{x}_{*}$.
In our case, the quantile of reward under no uncertainty of $\boldsymbol{w}$
is given by the inverse cumulative distribution function of beta distribution 
$F^{-1}(\cdot; \exp((\boldsymbol{\beta}+\boldsymbol{\rho})^\top\boldsymbol{x}_{*}), \exp(\boldsymbol{\rho}^\top\boldsymbol{x}_{*}))$. 
On the Monte Carlo samples 
$(\zeta_{*}^{(1)},\eta_{*}^{(1)}),\ldots,(\zeta_{*}^{(m)},\eta_{*}^{(m)})$ in (\ref{equation:monte_carlo_scores}),
the $\nu$-EUQ measure is empirically computed as
\begin{equation}
{EUQ}_{q(\theta_{*}\vert\boldsymbol{x}_{*}, {\mathcal D},\boldsymbol{c})}^{(\nu)}[\theta_{*}]\!=\!%
\frac{1}{m}\sum_{t=1}^mF^{-1}\!\left(\nu; %
\exp(\zeta_{*}^{(t)}+\eta_{*}^{(t)}), %
\exp(\zeta_{*}^{(t)})\right)\nonumber.%
\end{equation}

\paragraph{Upper Confidence Quantile of Upper Quantile (UCQUQ)}

A further optimistic statistic
is obtained through replacing the expectation in EUQ by upper quantile.
The $(\nu_1,\nu_2)$-UCQUQ measure is the $100\nu_2$-percentile of
the empirical posterior distribution about the $100\nu_1$-percentile of reward, as
\begin{eqnarray}
{UCQUQ}_{q(\theta_{*}\vert\boldsymbol{x}_{*}, {\mathcal D},\boldsymbol{c})}^{(\nu_1,\nu_2)}[\theta_{*}]\!&=&\!G_{m,\nu_1}^{-1}(\nu_2)\mbox{ and }\nonumber\\
G_{m,\nu_1}(\theta)\!&\triangleq&\!%
\frac{1}{m}\sum_{t=1}^m U\left(\theta-F^{-1}\!\left(\nu_1; %
\exp(\zeta_{*}^{(t)}+\eta_{*}^{(t)}), %
\exp(\zeta_{*}^{(t)})\right)\right)\nonumber,%
\end{eqnarray}
where $U(\cdot)$ is the unit step function
and $G_{m,\nu_1}^{-1}(\cdot)$ is the inverse function of $G_{m,\nu_1}(\cdot)$.

\paragraph{Upper Quantile of Predictive distribution (UQP)}

Predictive distribution can also supply an optimistic statistic
that has a similar principle to EUQ. Here the quantile is taken after the marginalization
over the posterior.
Upper quantile of the empirical predictive distribution is given as
\begin{equation}
\label{equation:upper_quantile_predictive}
{UQP}_{q(\theta_{*}\vert\boldsymbol{x}_{*}, {\mathcal D},\boldsymbol{c})}^{(\nu)}[\theta_{*}]\!=\!\theta_{\nu}\mbox{ such that }%
\nu=\frac{1}{m}\sum_{t=1}^m%
F\!\left(\theta_{\nu}; %
\exp(\zeta_{*}^{(t)}+\eta_{*}^{(t)}), %
\exp(\zeta_{*}^{(t)})\right).%
\end{equation}
While there is no closed-form formula of $\theta_{\nu}$
in (\ref{equation:upper_quantile_predictive}),
we can numerically compute the quantile
by applying bi-section method with the cumulative distribution function of Beta distribution
$F(\cdot; \cdot, \cdot)$.

\section{Related work}
\label{sec:related_work}

Beta-binomial-logit models have been used for robust classification
whereas their over-dispersion has been supplied by not a regression formula but by a scholar hyperparameter (e.g., \cite{tak2017data}).
Contextual-risk models have been used in regression tasks
such as Gaussian Process (GP) regression with input-dependent variances \cite{goldberg1998regression},
while have been uncommon in classification tasks.
Our derivation of the custom Laplace approximation and marginal likelihood
are based on the techniques in GP classification \cite{rasmussen2005gaussian},
for which Expectation Propagation (EP; \cite{minka2001expectation}) is also applicable whereas we avoided too complicated formulas of EP.
Nonparametric conditional density estimation (e.g., \cite{shahbaba2009nonlinear,sugiyama2010conditional})
naturally introduces input-dependent noises while simpler forms are preferable in our task.
$T$-logistic regression \cite{ding2010logistic} is another robust classifier and
Bayesian estimation of robust classifiers
produces robust credible intervals (e.g., \cite{pericchi1991robust}),
while their risks do not depend on inputs.
Overall, to the best of our knowledge, we provide the most parsimonious Bayesian classifier
that suits recommendation of outlying items based on input-dependent variabilities and uncertainties.

Humans exhibit systematically predictable behaviors in the face of uncertainty.
One reliable observation is that desire to avoid monetary loss is a strong incentive for exploration
(e.g., win-stay lose-shift algorithm \cite{robbins1952some}, prospect theory \cite{kahneman1979prospect},
loss aversion \cite{usher2004loss}, and regulatory focus theory
\cite{higgins2000making,avnet2006how}) and 
compensating money works as an incentive \cite{frazier2014incentivizing}.
Users do not lose money, however, in online news service when they read a narrow range of articles.
Even in mental level, it is uncertain whether reading merely one-sided opinions lets users feel pains.
Another observation is that diminishing return for the same type of stimulus
naturally leads exploration (e.g., \cite{smith1999diagnosing}, variety-seeking behavior in marketing \cite{kumar2006estimation,ho2014towards}) 
to maintain the Optimum Stimulation Level (OSL) \cite{raju1980optimum}.
Diminishing return has already been exploited in diversified recommendation (e.g., linear submodular bandits \cite{yue2011linear}),
while we already discussed the insufficient power of diversification for surely unpopular articles.

\section{Experimental evaluations}
\label{sec:experiments}

We experimentally evaluate the varieties of our scoring methods.
In Section \ref{subsec:performance_measure},
we define a performance indicator of early detectability for special articles 
that are chosen in the test period despite their dissimilarity to the positive training samples.
For real-world news-reading datasets and reference models introduced in Section \ref{subsec:experimental_setting},
we compare the performances among the proposed and reference models in Section \ref{subsec:performance_comparisons}.
The proposed models outperform the reference models in terms of the 
serendipity-oriented indicator, while achieving competitive levels of test-set likelihood.

\subsection{Serendipity-oriented performance measure}
\label{subsec:performance_measure}

Our main indicator is a variation of Area Under Curve
with prioritization of articles whose popularity is hard to be early detected.
Let $\boldsymbol{S}_i
\!\triangleq\!(S_{ijj'}\!\equiv\!\boldsymbol{x}_{ij}^\top\boldsymbol{x}_{ij'})\!\in\!{\mathbb R}^{N[i]\times N[i]}$ be
a  similarity matrix among all of the items that user $i$ chose in the training period, where
$N[i]\!\triangleq\!\vert\lbrace j; v_{ij}>0\wedge (i,j)\!\in\!{\mathcal D}\rbrace\vert$.
For test vector of context $\boldsymbol{x}_{i*}$,
we define an augmented $(N[i]\!+\!1)$-by-$(N[i]\!+\!1)$ matrix
$\boldsymbol{S}_{i*}$, via taking the inner product between the test vector
$\boldsymbol{x}_{i*}$ and every training vector assigned with a positive label.
By borrowing the common formula between the partition function in DPPs \cite{kulesza2013determinantal,kulesza2011learning}
and exponentiation of mutual information of GP \cite{contal2014gaussian},
we define reward variable $r_{i*}\!\geq\!1$ for this test context as
\begin{equation}
\label{diversity_gain}
r_{i*}\!\triangleq\!\dfrac{\det (\boldsymbol{I}_{N[i]+1}\!+\!\boldsymbol{S}_{i*})}{\det (\boldsymbol{I}_{N[i]}\!+\!\boldsymbol{S}_{i})}%
\!\equiv\!1\!+\!\boldsymbol{x}_{i*}^\top\boldsymbol{x}_{i*}%
\!-\!\left((\boldsymbol{X}_i\boldsymbol{x}_{i*})^\top%
(\boldsymbol{I}_{N[i]}+\boldsymbol{X}_{i}\boldsymbol{X}_i^\top)^{-1}%
(\boldsymbol{X}_i\boldsymbol{x}_{i*})%
\right)\!,%
\end{equation}
where $\boldsymbol{X}_i\!\in\!{\mathbb R}^{N[i]\times d}$ is user $i$-specific design matrix 
that lines up all of the positively-labeled context vectors.
Eq.~(\ref{diversity_gain}) represents a monotonically non-decreasing gain of diversity among the articles chosen by user $i$, 
when the test article is added into the existing selection. 
By setting each horizontal length $r_{i*}$ and unifying the test samples by all of the users,
we draw one Receiver Operator Characteristic curve.
We name the resulting performance measure the Serendipity-oriented Area Under Curve (SAUC).
While other reward variables are considerable,
we regard Eq.~(\ref{diversity_gain})
as a good starting point for further studies about serendipity,
because of its direction connection with diversity and entropy,

\subsection{Real-world news datasets and reference models}
\label{subsec:experimental_setting}

\begin{table}[t]
\centering
\caption{Basic characteristics of the four news datasets in 2016.
Each of the {\tt Brexit}, {\tt USElect}, and {\tt FBIMail} datasets covers one-week 
records of individual-level impressions and pageviews for selected users,
where a big political news occurred around the boundary between
the training and test periods. The {\tt Normal} dataset is introduced for 
evaluation when there is no big political news.}
\label{table:characteristics_datasets}
\centering
\begin{tabular}{r|r|r|r|r|r|r}
Dataset Name&Training Period&Test Period&\#users&\#items&$\sum_{(i,j)\in{\mathcal D}}v_{ij}$&$\sum_{(i,j)\in{\mathcal D}}n_{ij}$\\\hline
{\tt Brexit}&Jun 20-23&Jun 24-26&9,908&9,026&221,781&2,388,617\\\hline
{\tt USElect}&Nov 5-8&Nov 9-11&9,841&2,419&266,331&2,792,707\\\hline
{\tt FBIMail}&Aug 29-Sep 1&Sep 2-4&9,871&2,123&184,492&2,060,274\\\hline
{\tt Normal}&Apr 1-4&Apr 5-7&9,896&6,618&186,557&1,995,538\\\hline
\end{tabular}
\end{table}

Our news datasets consist of individual-user-level 
page view and impression logs of a mobile-app news service in the United States.
Table \ref{table:characteristics_datasets} shows the characteristics of our 4 datasets.

The vector of context $\boldsymbol{x}_{ij}$ is given by text content of article
and clustering of users.
For each user $i$, we have a binary vector 
to indicate the items that user $i$ chose before the training period.
With standardizing the $L_2$-norm of every binary vector,
we performed clustering of users by spherical $64$-means algorithm.
We assume that cluster label represents each user's preference
that does not change during every one-week period.
Let $\boldsymbol{x}_j\!\in\!{\mathbb R}^{d_0}$ be unit-norm TF-IDF vector 
of article $j$. 
For multi-task learning \cite{chapelle2014simple}, we set
$\boldsymbol{x}_{ij}\!=\!%
(\boldsymbol{x}_j^\top,\boldsymbol{0}_{d_0}^\top,\ldots,\boldsymbol{0}_{d_0}^\top,\boldsymbol{x}_j^\top,%
\boldsymbol{0}_{d_0}^\top,\ldots,\boldsymbol{0}_{d_0}^\top)^\top\!\in\!{\mathbb R}^{(64\!+\!1)d_0}$ where the positions of non-zero elements are determined based on the cluster label of user $i$.
This feature design is used for all of the models.
Design of more sophisticated feature vector is out of this paper's scope.

Because our model's structural advantage comes from the context-dependence of variabilities and uncertainties,
the reference methods are {\tt M-Log}: ordinary $L_2$-regularized MAP logistic regression,
{\tt M-BBL}: MAP estimate of a Beta-binomial-logit model with input-independent over-dispersion \cite{tak2017data},
{\tt L-Log} and  {\tt L-BBL}: Bayesian extensions of {\tt M-Log} and  {\tt M-BBL} by Laplace's method, respectively.
{\tt L-Log} is a linear GP classifier and see \cite{rasmussen2005gaussian} for the concrete formulas.
{\tt M-BBL} and {\tt L-BBL} are the estimates when we replace $\boldsymbol{\rho}\!^\top\!\boldsymbol{x}_{ij}$ in (\ref{equation:beta_model}) by scholar parameter $\rho_0$.
The proposed model and its MAP counterpart are named {\tt L-Prop} and {\tt M-Prop}, respectively.
All of the models are fitted with empirical-Bayes method,
where every $L_2$-regularization hyperparameter $c_{(\cdot)}$
is optimized by gradient descent with initialization such that 
$c_{(\cdot)}\!:=\!\sum_{(i,j)\in{\mathcal D}}n_{ij}$.

\subsection{Performance comparisons}
\label{subsec:performance_comparisons}

\begin{figure*}[t]
  \centering
\begin{minipage}{0.495\textwidth}%
  \centering
  \includegraphics[width=0.95\textwidth]{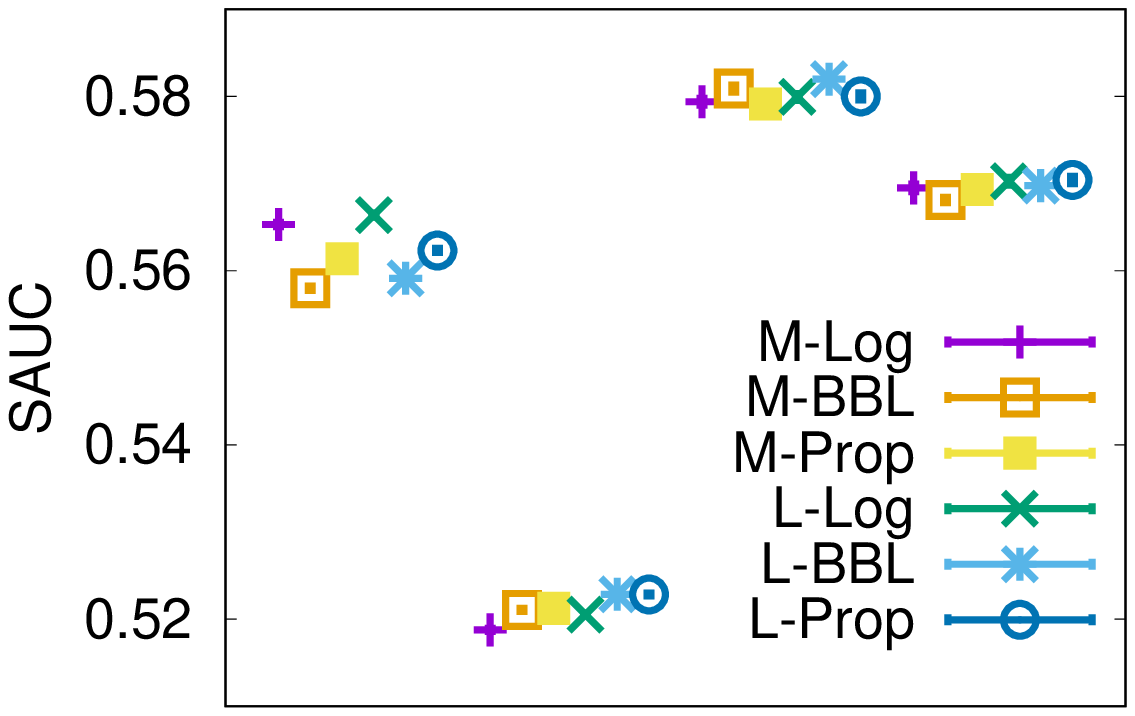}\\
\vspace*{-14pt}%
SAUC by MAP estimate or 0.95-UCQE
\vspace*{14pt}%
\end{minipage}%
\begin{minipage}{0.495\textwidth}%
  \centering
  \includegraphics[width=0.95\textwidth]{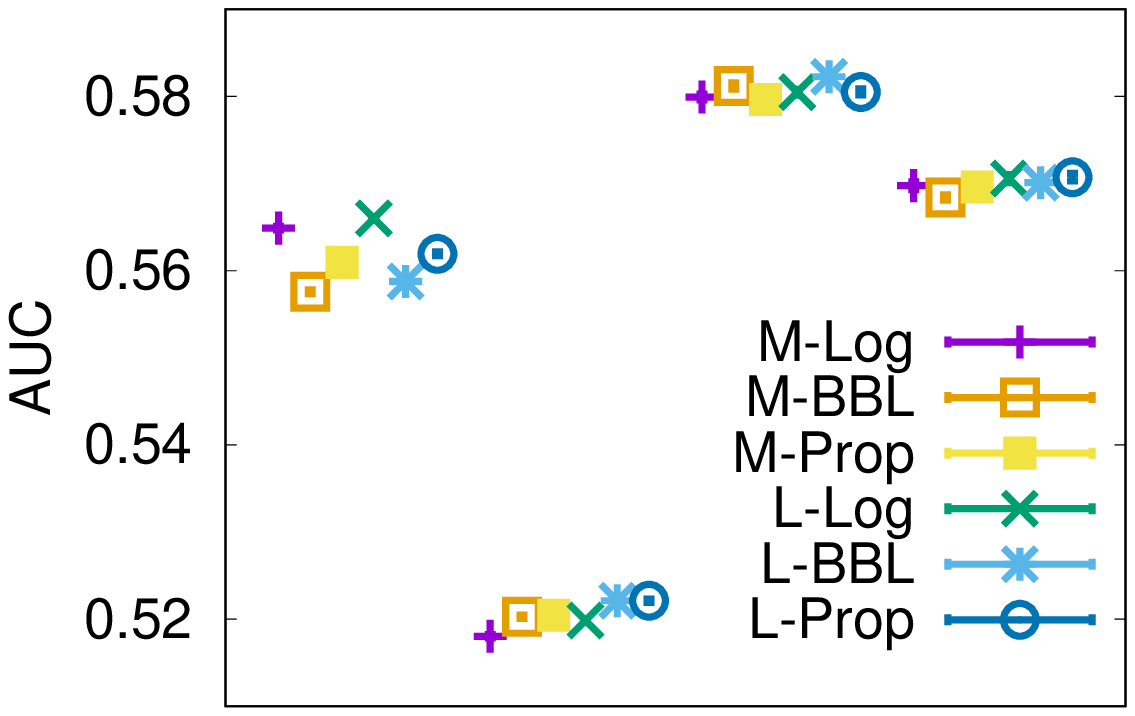}\\
\vspace*{-14pt}%
AUC by MAP estimate or 0.95-UCQE
\vspace*{14pt}%
\end{minipage}\\
\centering
\begin{minipage}{0.495\textwidth}%
  \centering
  \includegraphics[width=0.95\textwidth]{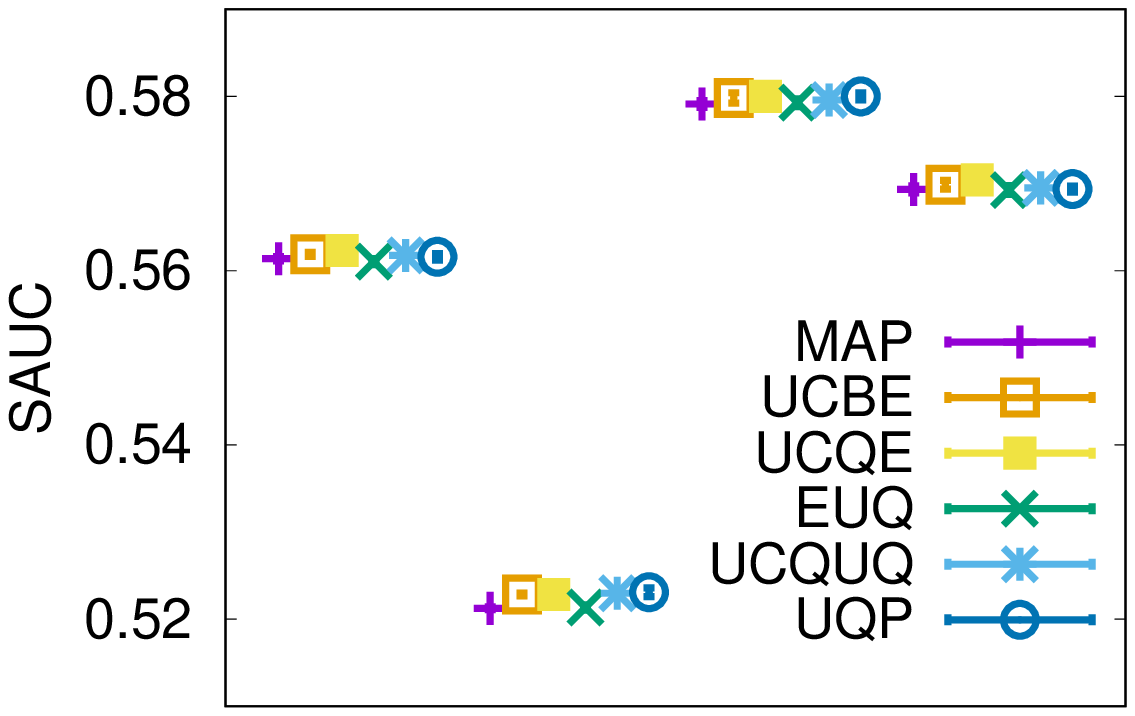}\\
\vspace*{-14pt}%
SAUC by various scoring\\
within the proposed model
\vspace*{14pt}%
\end{minipage}%
\begin{minipage}{0.495\textwidth}%
  \centering
  \includegraphics[width=0.95\textwidth]{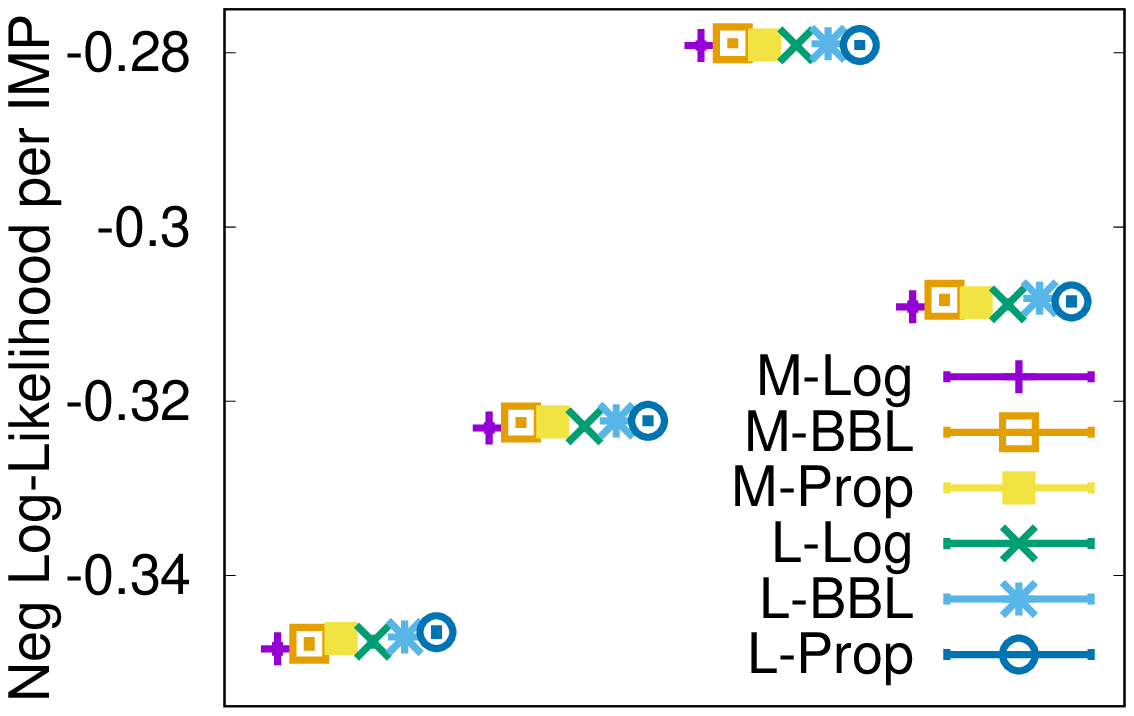}\\
\vspace*{-14pt}%
Average log-likelihood\\per impression
\vspace*{14pt}%
\end{minipage}%
\caption{Performances measures when $\nu_1\!\equiv\!\nu_2\!\equiv\!\nu\!\equiv\!0.95$.
Every score is 5-fold bootstrap mean for test dataset,
where the deviation among the 5 folds was found to be negligibly small.
For fair comparison, the test likelihood for the Beta-binomial models
is evaluated independently for each impression, while the training likelihood is not independent.
The proposed model {\tt L-Prop} achieved the highest likelihood for all of the 4 datasets
while its SAUC and AUC are highest only for 2 datasets {\tt USElect} and {\tt Normal}.
Among the scoring methods, expectation-oriented
UCBE and UCQE measures or predictive-distribution-based UQP are shown to be promising.
}
\label{figure:performance}
\end{figure*}

Figure \ref{figure:performance} exhibits several performance comparisons
by using SAUC and test-set log-likelihood per impression. AUC is also shown
for comparison. The proposed model achieved the highest likelihood for all of the 4 datasets
and highest SAUC for 2 datasets.
While the differences among models and scoring measures are currently marginal,
the advantage of the proposed model is significant
because of the negligibly small standard deviation among the 5 bootstrap folds.

\section{Conclusion}
\label{sec:conclusion}
This paper introduced a new contextual Bayesian binary classifier
whose risks and/or uncertainties on probabilities of success have been underestimated by existing models.
The proposed model consists of a conditional Beta distribution on input vector of context,
and its estimation is done with its low-rank Laplace approximation using thin SVD.
The closed forms for our approximate posterior yield several uncertainty-aware scoring measures,
and parts of them were experimentally successful in early detecting the exceptional articles
that are clicked despite the dissimilarity to the positive examples in training period.

In the future, we will apply our philosophy on context-dependent risks and uncertainties
for structured non-linear models. We will also qualitatively investigate
article examples associated with high uncertainties.
We hope our work to become a catalyst for further studies in computational journalism.

\newpage
\small{
\bibliographystyle{unsrt}
\bibliography{anti_filter_bubble}
}
\end{document}